\documentclass[11pt]{article}

\usepackage[preprint]{acl}

\usepackage{times}
\usepackage{latexsym}

\usepackage[T1]{fontenc}

\usepackage[utf8]{inputenc}

\usepackage{microtype}

\usepackage{inconsolata}

\usepackage{graphicx}
\usepackage{textcomp}
\usepackage{makecell}
\usepackage{booktabs}
\usepackage{multirow}
\usepackage[table]{xcolor}
\usepackage{amsmath}
\usepackage{amssymb} 
\usepackage[most]{tcolorbox}
\usepackage{paralist}
\usepackage{enumitem}
\usepackage{tabularx}
\usepackage{pifont}
\usepackage[ruled,vlined,linesnumbered]{algorithm2e}

\makeatletter
\DeclareRobustCommand\onedot{\futurelet\@let@token\@onedot}
\def\@onedot{\ifx\@let@token.\else.\null\fi\xspace}

\makeatother
\newcommand{\Tref}[1]{Table~\ref{#1}}

\newcommand{\Fref}[1]{Figure~\ref{#1}}

\usepackage{xspace}
\newcommand{\modelname}{\textsc{IceBreaker}\xspace}
\definecolor{myblue}{HTML}{c2d6eb}
\definecolor{mylightblue}{HTML}{f0f3fb}

\usepackage[normalem]{ulem}

\newcommand{\iceemoji}[1][1.2em]{\raisebox{-0.15em}{\includegraphics[height=#1]{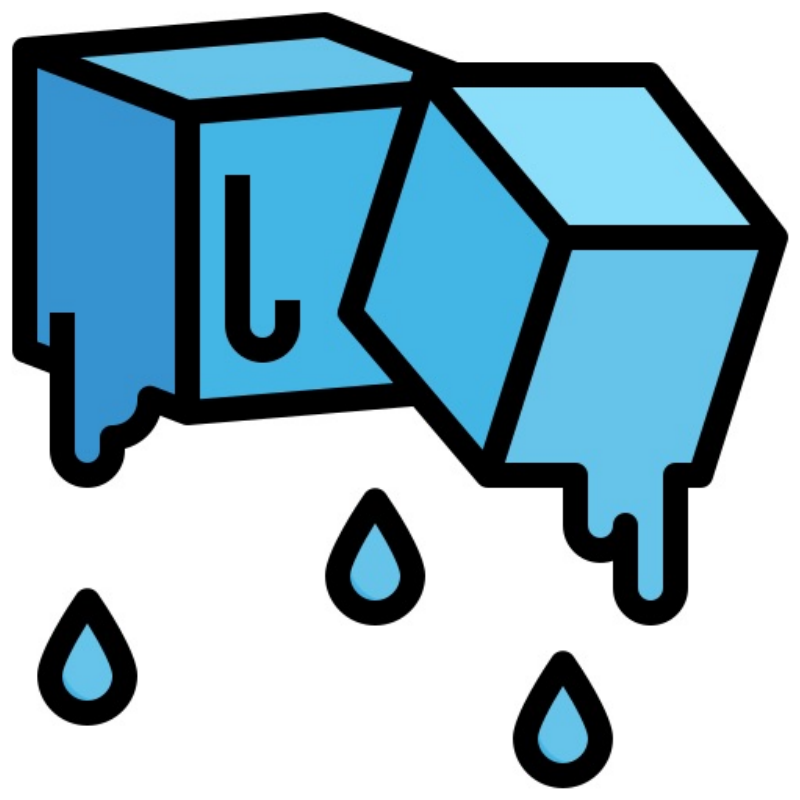}}}
\newcommand{\cbox}[2]{\colorbox{#1}{\raisebox{0pt}[1.2ex][0.1ex]{#2}}}
\usepackage{soul}
\SetAlCapFnt{\small}
\SetAlCapNameFnt{\small\bfseries}

\newcommand{\bemail}[1]{\href{mailto:#1@bytedance.com}{\texttt{#1}}}

\title{\iceemoji \;\modelname for Conversational Agents:\\Breaking the First-Message Barrier with Personalized Starters}

\author{
 \textbf{Hongwei Zheng}\thanks{Equal contribution.},
 \textbf{Weiqi Wu}\footnotemark[1],
 \textbf{Zhengjia Wang}\footnotemark[1],
 \textbf{Guanyu Jiang},\\
 \textbf{Haoming Li},
 \textbf{Tianyu Wu},
 \textbf{Yongchun Zhu}\thanks{Corresponding author.},
 \textbf{Jingwu Chen}\footnotemark[2],
 \textbf{Feng Zhang}
\\
 ByteDance
\\
 \small{\{\bemail{zhenghongwei}, \bemail{wuweiqi}, \bemail{wangzhengjia.jia}, \bemail{jiangguanyu.jgy},}
 \\
 \small{\bemail{lihaoming.cs}, \bemail{wutianyu.23}, \bemail{zhuyongchun.zyc}, \bemail{chenjingwu}, \bemail{feng.zhang}\}@bytedance.com}
}

\begin{document}
\maketitle

\begin{abstract}

Conversational agents, such as ChatGPT and Doubao, have become essential daily assistants for billions of users.
To further enhance engagement, these systems are evolving from passive responders to proactive companions.
However, existing efforts focus on activation within ongoing dialogues, while overlooking a key real-world bottleneck. 
In the conversation initiation stage, users may have a vague need but no explicit query intent, creating a first-message barrier where the conversation holds before it begins.
To overcome this, we introduce \textbf{Conversation Starter Generation}: generating personalized starters to guide users into conversation. 
However, unlike in-conversation stages where immediate context guides the response, initiation must operate in a cold-start moment without explicit user intent.
To pioneer in this direction, we present \textbf{\modelname} that frames human ice-breaking as a two-step handshake: \textit{(i) evoke resonance} via Resonance-Aware Interest Distillation from session summaries to capture trigger interests, and \textit{(ii) stimulate interaction} via Interaction-Oriented Starter Generation, optimized with personalized preference alignment and a self-reinforced loop to maximize engagement.
% ToChange
Online A/B tests on one of the world's largest conversational agent products show that \modelname improves user active days by +1.84\textperthousand\ and click-through rate by +94.25\textperthousand, and has been deployed in production.
\end{abstract}

\section{Introduction}

\begin{figure}[t]
\centering
  \includegraphics[width=\linewidth]{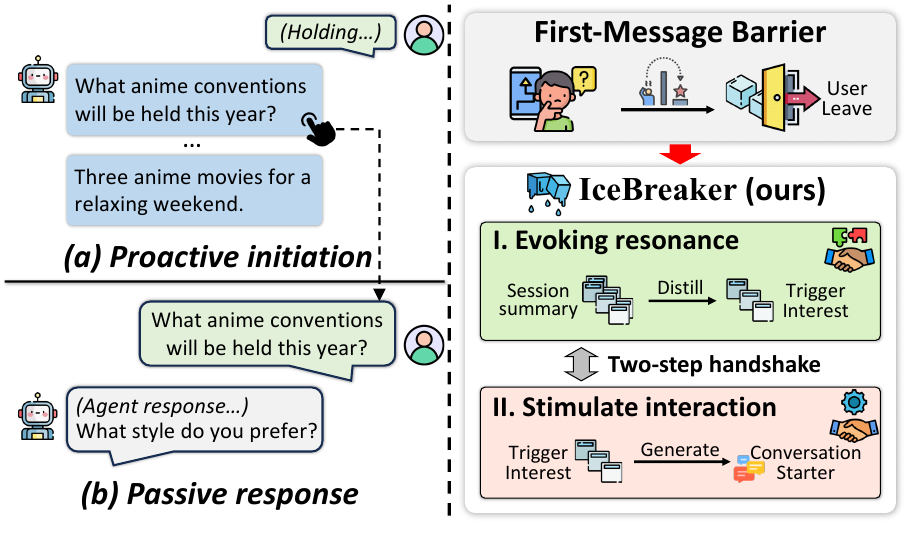}
  \setlength\abovecaptionskip{-8pt}
  \setlength\belowcaptionskip{-10pt}
  \caption{
  Paradigm comparison between (a) proactive initiation that lowers the first-message barrier by providing personalized \cbox{myblue}{starters} and (b) passive \cbox{gray!25}{response} that occurs within an ongoing dialogue. \textbf{\modelname} mimics human ice-breaking via a two-step handshake: evoking resonance and stimulating interaction.
  }
  \label{fig:motiv}
  % \vspace{-1em}
\end{figure}

Conversational agents are rapidly becoming an indispensable part of everyday life \citep{hadi2023survey}, people turn to them not only for information seeking \citep{zhu2025large} and task assistance \citep{qin2024tool}, but also for planning and emotional companionship \citep{kasneci2023chatgpt, li2023systematic}.
In this work, we study a deployment setting on one of the world's largest conversational agent products, serving hundreds of millions of users.
In real-world deployments, the serving paradigm is shifting from passive answering to proactive participation, where the system anticipates needs and helps users move the interaction forward \citep{deng2025proactive}. Prior studies on such \textit{\textbf{in-conversation}} proactivity operate after the user has already started the conversation (as depicted in \Fref{fig:motiv}(b)), for example by generating follow-up or clarifying questions \citep{deng2023prompting}, and have been shown to improve session depth \citep{andukuri2024stargate, li2025proactive}.

In practice, however, a fundamental product bottleneck remains under-explored: the first-message barrier in the \textit{\textbf{before-conversation}} stage.
At this cold-start moment, users often have only a vague goal and a narrow sense of the agent's actionable scope \citep{norman2013design, zamfirescu2023johnny}, making it difficult to start the conversation and highlighting the need for proactive guidance.
Production statistics also show that this ``\textbf{\textit{first-message barrier}}'' correlates with a lower conversation start rate: \textit{roughly 20\% of users enter the product but leave without starting the conversation}.
To address this bottleneck, we study \textbf{proactive initiation}, which provides personalized guidance that lowers the cost of conversation initiation, makes the agent's capabilities immediately actionable, and more importantly, shapes the subsequent trajectory to improve downstream conversation quality.

However, proactive initiation poses challenges beyond standard response generation. 
First, in the before-conversation stage, inferring user intent without explicit user signals is challenging. 
Second, user preferences are highly personalized and long-tailed, so a one-size-fits-all alignment objective tends to bias generation toward generic starters that fail to resonate with individuals.
To pioneer in this direction, we study proactive initiation as the task of \textbf{Conversation Starter Generation}: producing a set of personalized starter questions that guide users to start a conversation. 
Motivated by how people initiate conversations in cold-start situations, first surfacing a few interests likely to resonate and then phrasing a starter to elicit interaction, we propose \textbf{\modelname}, which operationalizes proactive initiation as a two-step handshake: 
\textit{(i) evoke resonance}, where Resonance-Aware Interest Distillation (RID) learns a personalized resonance scorer to distill long session summaries into a compact set of trigger interests, with activity-aware gating controlling distillation strength; and
\textit{(ii) stimulate interaction}, where Interaction-Oriented Starter Generation (ISG) conditions on distilled interests to generate a small, diverse starter list, warm-started with distilled instruction data and then aligned via list-wise multi-dimensional preference optimization with periodically augmented preference pairs for self-reinforced refinement.

% ToChange
Online A/B tests show a +1.84\textperthousand\ user active days increase and +94.25\textperthousand\ CTR improvement. \modelname has been deployed at scale on one of the world's largest conversational agent products.
Our main contributions include:
\begin{itemize}[leftmargin=*,itemsep=2pt,topsep=0pt,parsep=0pt]
    \item \textbf{Paradigm shift:} By formalizing Conversation Starter Generation, we move from responsive execution to proactive initiation to address the first-message barrier in real-world applications.
    
    \item \textbf{\modelname:} We propose a two-step handshake framework that bridges the resonance gap in cold-start initiation by coupling resonance-aware interest distillation with preference-aligned starter generation.
    
    \item \textbf{Real-world deployment:} Extensive offline and online experiments demonstrate that \modelname not only improves conversation initiation experience but also increases long-term engagement. \modelname has been deployed at scale, serving hundreds of millions of users.
    
\end{itemize}

\begin{figure*}[t]
  \includegraphics[width=\linewidth]{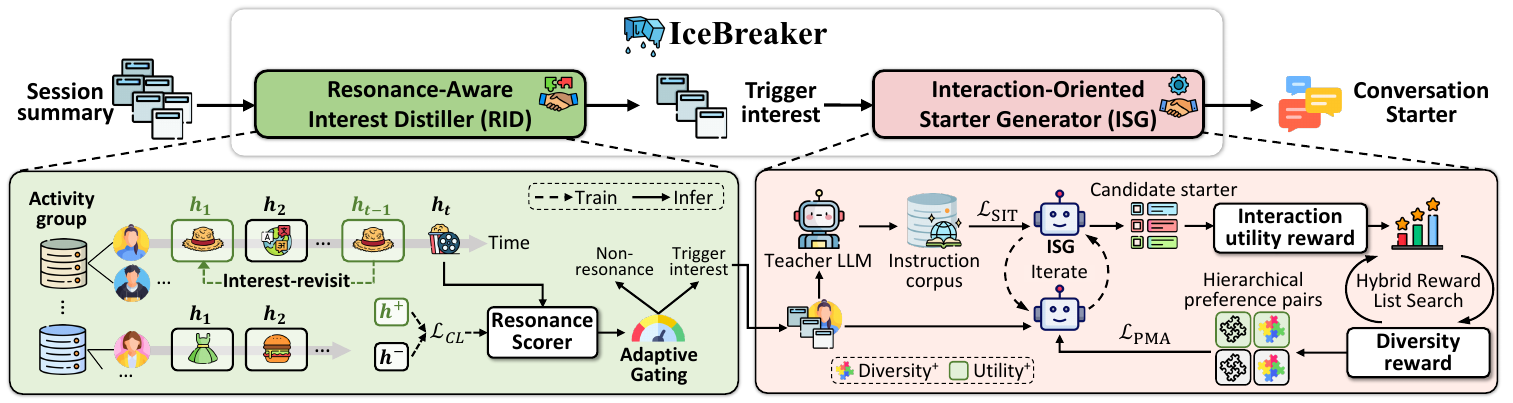}
  \setlength\abovecaptionskip{-8pt}
  \setlength\belowcaptionskip{-6pt}
  \caption{Overall architecture of \textbf{\modelname}. \textbf{(a) Resonance-Aware Interest Distillation} identifies trigger interests from session summaries via personalized resonance scoring and adaptive gating. \textbf{(b) Interaction-Oriented Starter Generation} produces a small set of first-turn starters conditioned on these interests, and optimizes them for personalized interaction utility and within-list diversity through preference alignment.}
  \label{fig:method}
  % \vspace{-1em}
\end{figure*}

\section{Proposed Method: \modelname}
\label{sec:methodology}

In this section, we present \modelname, a two-stage framework for proactive initiation under cold-start moments.
As illustrated in \figurename~\ref{fig:method}, \modelname consists of a distiller that captures trigger interests from session summaries (\S~\ref{sec:interest}), and a generator that produces a list of conversation starters conditioned on these interests (\S~\ref{sec:generation}).

\subsection{Problem Formulation}
\label{sec:problem_formulation}
We first formalize the \textbf{Conversation Starter Generation} task.
Given a user’s session summaries, the goal is to generate a small list of starters that maximize subsequent engagement.
Let $\mathcal{H}_u = \{h_1, h_2, \dots, h_T\}$ denote the historical session summaries of user $u$, where each $h_t$ is a short textual summary of the $t$-th session.
The goal is to learn a policy $\pi_\theta$ that generates a list of $K$ conversation starters $\mathcal{S} = [s_1, s_2, \dots, s_K]$.
The objective is to maximize the expected list-wise utility:
\begin{equation}
    \mathcal{S}^* = \arg\max_{\mathcal{S}} \mathbb{E}_{\mathcal{S} \sim \pi_\theta(\cdot|\mathcal{H}_u)} \mathrm{V}(\mathcal{S}),
\end{equation}
where $\mathrm{V}(\mathcal{S})$ denotes the overall value of the list, which depends on both its user-specific utility and the redundancy within the list. 
In \modelname, this objective is approximated with a learned interaction-utility ranker and an explicit within-list diversity signal (\S~\ref{sec:generation}).

\subsection{Resonance-Aware Interest Distillation (RID)} \label{sec:interest}
As the first handshake for resonance evoking, RID captures trigger interests from session summaries by mimicking how humans initiate conversations: instead of enumerating all possible topics, people surface a small set of personal resonance cues that are most likely to elicit interaction.
We implement RID as a resonance-guided distillation pipeline: we first learn a personalized resonance scorer from user actions, and then conduct adaptive gating to produce trigger interests $\mathcal{I}^*$ from $\mathcal{H}_u$.

\paragraph{Personalized Resonance Scorer}
To quantify user-specific resonance potential, we learn a personalized resonance scorer $\text{s}_{\phi}(\cdot)$ that maps a candidate session summary $h_t$ of user $u$ to a scalar score.
Following \citet{deng2025onerec} and \citet{guo2025onesug}, the historical session summaries together with user profile (a crowd portrait fitted by the platform) are encoded as user features $\mathbf{u}=\text{E}_\text{user}(\mathcal{H}_u, u)$, session summary is encoded as $\mathbf{z}_t=\text{E}_\text{text}(h_t)$:
\begin{equation}
    \text{s}_{\phi}(u,h_t)=\cos(\mathbf{u},\mathbf{z}_t),
\end{equation}
where $\cos(\cdot,\cdot)$ denotes cosine similarity between $\ell_2$-normalized embeddings. This design enables efficient retrieval over large summary pools, which is important for real-world production.

For $\text{s}_{\phi}(\cdot)$ training, an \textit{interest-revisit} signal is leveraged as a proxy for resonance, which provides direct evidence of what a user is willing to revisit. 
Concretely, we identify resonance by matching a user's later session summaries to their historical session summaries using a prompting-based approach (detailed in \appendixname~\ref{appendix:prompt}). A session summary is considered as positive $h^{+}$ if a later session revisits the same interest. 
Two types of negatives are constructed: (i) within-user negatives $\mathcal{N}_{\text{intra}}$, i.e., other session summaries from the same user that are not matched by any later session, and (ii) cross-user negatives $\mathcal{N}_{\text{inter}}$, i.e., session summaries from other users in the same minibatch. Finally, $\mathcal{N}_u=\mathcal{N}_{\text{intra}} \cup \mathcal{N}_{\text{inter}}$ is used as the negative set.

Let $\mathcal{D}_{\text{RID}}$ denote the resulting set of training tuples $(u,h^{+},\mathcal{N}_u)$. We train the encoders in $\text{s}_{\phi}$ by minimizing the following contrastive loss:
\begin{equation}
\begin{aligned}
    \mathcal{L}_{\text{CL}}(\phi)
    &=
    - \frac{1}{|\mathcal{D}_{\text{RID}}|}
    \sum_{\mathcal{D}_{\text{RID}}}
    \log
    \frac{\exp(\text{s}_{\phi}(u,h^{+}))}
    {\sum_{h}\exp(\text{s}_{\phi}(u,h))},
\end{aligned}
\end{equation}
where $h \in \{h^{+}\}\cup \mathcal{N}_u$, $\phi$ denotes the parameters of scorer $\text{s}_{\phi}$, $|\mathcal{D}_{\text{RID}}|$ is the number of training tuples.

\paragraph{Adaptive Gating}
At inference, we compute scores $\{\text{s}_{\phi}(u,h_t)\}_{t=1}^{T}$ and distill trigger interests via an user activity-aware threshold.
Concretely, we map each user to an activity group $b(u)\in\{1,\dots,B\}$ based on user activity, and apply a group-specific threshold $\tau_{b}$ that increases with activity.
Formally, $\tau_u=\tau_{b(u)}$, where $\tau_{b}$ represents the learned score threshold for group $b$:
\begin{equation}
    \mathcal{I}^{*}=\{h_t \in \mathcal{H}_u \mid \text{s}_{\phi}(u,h_t)\ge \tau_u\}.
\end{equation}
The resulting $\mathcal{I}^{*}$ serves as resonance-aware cues for the downstream starter generator.

\subsection{Interaction-Oriented Starter Generator (ISG)} \label{sec:generation}
As the second handshake for stimulating reaction, conditioned on the distilled trigger interests $\mathcal{I}^*$, ISG generates a small list of first-turn starters $\mathcal{S}=[s_1,\dots,s_K]$ optimized for both personalized interaction utility and within-list diversity.

We train ISG in two stages: (i) Supervised Interest-Expanded Instruction Tuning (SIT) to warm-start generation quality and topical coverage, and (ii) Personalized Multi-Dimensional Alignment (PMA) to align list-level utility and diversity with user-specific feedback.

\subsubsection{Supervised Interest-Expanded Instruction Tuning} \label{sec:SIT}
We warm-start the generator using an interest--starter instruction corpus distilled from a teacher LLM.
Given trigger interests $\mathcal{I}^*$, the teacher generates qualified starters under multiple constraints, yielding a coverage-oriented instruction corpus $\mathcal{D}_{\text{cov}} = \{(\mathcal{I}^*, \mathcal{S})\}$.
Let $\pi_{\theta}(\cdot|\mathcal{I}^*)$ denotes the conversation starter generator distribution. The instruction-tuning objective is:
\begin{equation}
    \mathcal{L}_{\text{SIT}}(\theta) = - \mathbb{E}_{(\mathcal{I}^*, \mathcal{S}) \sim \mathcal{D}_{\text{cov}}} \left[ \log \pi_{\theta}(\mathcal{S}|\mathcal{I}^*) \right].
\end{equation}
SIT expands the training distribution beyond the limited observation and provides a stable initialization for subsequent preference optimization.

\begin{algorithm}[tb]
\small
  \setlength\abovecaptionskip{4pt}
  \setlength\belowcaptionskip{-8pt}
\SetAlgoLined
\caption{\small Hybrid Reward List Search}
\label{alg1}
\KwIn{user $u$, trigger interests $\mathcal{I}^*$, generator $\pi_{\theta}$, $\mathrm{R}_\text{util}(u,\cdot)$, $\mathrm{R}_\text{div}(\cdot,\cdot)$, list length $K$, pool size $N$, beam width $J$, expansion size $M$, diversity weight $\lambda$}
Iteratively sample the candidate pool from: $\mathcal{C} \gets \textsc{Sample}\big(\pi_{\theta}(\cdot|\mathcal{I}^*), N\big)$\;
Initialize: $\mathcal{B} \gets \{\langle\rangle\}$\;$\mathrm{V}(\langle\rangle)\gets 0$\;
\For{$\ell \gets 1$ \KwTo $K$}{
  Expand beam with utility and diversity reward\;
  $\mathcal{B}' \gets \emptyset$\;
  \ForEach{$\mathcal{P} \in \mathcal{B}$}{
    $\mathcal{T} \gets \textsc{Top}\big(\mathcal{C}\setminus\mathcal{P}, \mathrm{R}_\text{util}(u,\cdot), M\big)$\;
    \ForEach{$s \in \mathcal{T}$}{ 
      $\mathcal{P}' \gets \mathcal{P} \oplus s$\;
      $\mathrm{V}(\mathcal{P}') \gets \mathrm{V}(\mathcal{P}) + \mathrm{R}_\text{util}(u,s) + \lambda\cdot \mathrm{R}_{\text{div}}(\mathcal{P}, s)$\;
    }
  }
  Keep top-$J$ partial lists: $\mathcal{B} \gets \textsc{Top}\big(\mathcal{B}', \mathrm{V}(\cdot), J\big)$\;
}
$\mathcal{S}^{+} \gets \arg\max_{\mathcal{S}\in \mathcal{B}} \mathrm{Score}(\mathcal{S})$ 

$\mathcal{S}^{-}_{\text{util}}, \mathcal{S}^{-}_{\text{div}}, \mathcal{S}^{-}_{\text{jf}} \gets \textsc{SelectNegatives}(\mathcal{B}, \mathcal{S}^{+})$ 

\KwOut{$\mathcal{S}^{+}$, $\mathcal{S}^{-}_{\text{util}}, \mathcal{S}^{-}_{\text{div}}, \mathcal{S}^{-}_{\text{jf}}$}
\end{algorithm}
% \vspace{-14pt}

\subsubsection{Personalized Multi-Dimensional Alignment}
SIT teaches the generator to produce well-formed starters, but it does not align generation with user-specific interaction utility.
To align the generator with personalized feedback under sparse and evolving signals, PMA optimizes the generator with Direct Preference Optimization (DPO) and mines user-conditioned preference supervision tailored to both interaction utility and within-list diversity:
\begin{equation}
\begin{aligned}
    \mathcal{L}_{\text{PMA}}(\theta)
    &=
    - \mathbb{E}_{(\mathcal{S}_w, \mathcal{S}_l)\sim \mathcal{P}}
    \Big[
    \log \sigma \Big(
    \beta \big(
    \log \pi_{\theta}(\mathcal{S}_w|\mathcal{I}^*)
    \\
    &\qquad\qquad
    -
    \log \pi_{\theta}(\mathcal{S}_l|\mathcal{I}^*)
    \big)
    \Big)
    \Big],
\end{aligned}
\end{equation}
where $\beta$ is a temperature parameter and $\mathcal{S}_w \succ \mathcal{S}_l$ denotes list $\mathcal{S}_w$ is preferred to $\mathcal{S}_l$ for a given user.

\paragraph{Hybrid Reward List Search}
To obtain such user-conditioned preference pairs, a straightforward approach is to derive them from user feedback data. However, feedback from a specific user is extremely sparse, resulting in an insufficient candidate space that limits both the quantity and diversity of preference supervision.
To alleviate this personalized-feedback sparsity, we iteratively construct user-conditioned preference pairs from model-generated candidates during PMA.

Specifically, as shown in Algorithm~\ref{alg1}, we sample a candidate pool $\mathcal{C}$ from the current generator $\pi_{\theta}(\cdot|\mathcal{I}^*)$ for each user. Guided by a hybrid reward signal (interaction utility reward and diversity reward; denoted as $\mathrm{R}_\text{util}$ and $\mathrm{R}_\text{div}$ and detailed below), the algorithm incrementally grows partial lists: at each position, it expands a beam of partial lists with top-$M$ utility-ranked candidates and re-scores each expansion by combining utility with marginal diversity gain. As $\pi_{\theta}$ is updated across iterative preference alignment rounds, we re-sample $\mathcal{C}$ from the latest $\pi_{\theta}$ to mine fresh preference pairs.

\paragraph{Utility--Diversity Reward}
We guide list search and preference mining with a hybrid reward signal that jointly captures two dimensions:
\begin{itemize}[leftmargin=*,itemsep=2pt,topsep=0pt,parsep=0pt]
    \item Interaction utility reward $\mathrm{R}_\text{util}(u,s)$: the personalized likelihood that starter $s$ triggers interaction for user $u$.
    \item Diversity reward $\mathrm{R}_\text{div}(\mathcal{P},s)$: the marginal diversity gain of adding $s$ to a partial list $\mathcal{P}$, computed from sentence embeddings.
\end{itemize}
Algorithm~\ref{alg1} maintains a running list value $\mathrm{V}(\mathcal{P})$ by accumulating these stepwise rewards.
For interaction utility, $\mathrm{R}_\text{util}(u,s)$ captures both short-term and long-term value: (i) predicting whether the user engages with the presented starter, and (ii) predicting the depth of subsequent conversations measured by the number of sessions. The model outputs a point-wise score in $(0,1)$:
\begin{equation}
    \mathrm{R}_\text{util}(u,s)=\text{Proj}\big([\text{E}_{\text{user}}(\mathcal{H}_u, u);\text{E}_{\text{text}}(s)]\big),
\end{equation}
where $\text{Proj}(\cdot)$ is the projection layer and $\text{E}(\cdot)$ denotes the text encoder.
For a completed list $\mathcal{S}=[s_1,\dots,s_K]$, we compute its normalized utility $\hat{v}_{\text{util}}(\mathcal{S})=\frac{1}{K}\sum_{k=1}^{K}\mathrm{R}_\text{util}(u,s_k)$ and its diversity $\hat{v}_{\text{div}}(\mathcal{S})\in[0,1]$ as the average pairwise cosine dissimilarity among sentence embeddings. We then select the preferred list by:
\begin{equation}
\begin{aligned}
    &\mathcal{S}^{+}=\arg\max_{\mathcal{S}} \mathrm{V}(\mathcal{S}), \\& \mathrm{V}(\mathcal{S})=\hat{v}_{\text{util}}(\mathcal{S})+\lambda\cdot \hat{v}_{\text{div}}(\mathcal{S}),
\end{aligned}    
\end{equation}
where $\lambda$ controls the utility--diversity trade-off.

Three types of dispreferred lists are constructed for hierarchical supervision:
\begin{itemize}[leftmargin=*,itemsep=2pt,topsep=0pt,parsep=0pt]
    \item Utility negatives ($\mathcal{S}^{-}_\text{util}$): lists with substantially lower $\hat{v}_{\text{util}}$ than $\mathcal{S}^{+}$ while keeping $\hat{v}_{\text{div}}$ comparable, isolating the signal for user-specific utility.
    \item Diversity negatives ($\mathcal{S}^{-}_\text{div}$): lists with substantially lower $\hat{v}_{\text{div}}$ while keeping $\hat{v}_{\text{util}}$ comparable, providing the signal for within-list non-redundancy.
    \item Joint-failure negatives ($\mathcal{S}^{-}_\text{jf}$): lists that are poor on both $\hat{v}_{\text{util}}$ and $\hat{v}_{\text{div}}$, providing easy negatives that stabilize preference learning.
\end{itemize}
Finally, $\mathcal{S}^{-} \in \{\mathcal{S}^{-}_\text{util}, \mathcal{S}^{-}_\text{div}, \mathcal{S}^{-}_\text{jf}\}$, which disentangle utility and diversity trade-offs for DPO training, enabling the generator to improve engagement without collapsing into generic or repetitive starters.

\paragraph{Self-Reinforced Iterative Optimization}

To continuously harvest informative supervision under sparse feedback, we alternate between preference mining and policy optimization by iteratively mining hierarchical preference pairs from the latest $\pi_{\theta}$, augmenting them with teacher-distilled starters, and updating $\pi_{\theta}$ on the accumulated preference set.
In deployment, we periodically execute this procedure to track drifting user preferences.

\section{Experiments}
\label{sec:experiments}

We evaluate \modelname with offline benchmarks and online A/B tests, and conduct ablations to quantify the contribution of each component.

\subsection{Experimental Setup}

\paragraph{Datasets}
We construct the offline benchmark from anonymized user actions collected in a large-scale conversational system, using activity stratification to ensure balanced representation across engagement levels.
For each user, historical session summaries are retrieved to generate candidate starters, which are then evaluated with the interaction utility ranker.
Final effectiveness is validated through online A/B tests for more than one month.

\paragraph{Baselines}
We compare \modelname against two categories of LLM-based generation.
For training-free baselines:
$\bullet$ \textbf{PE} directly prompts the backbone model with session summaries to generate starters;
$\bullet$ \textbf{PE + RID} prompts the model with distilled trigger interests by Resonance-Aware Interest Distillation.
For fine-tuned baselines:
$\bullet$ \textbf{SFT} fine-tunes the starter generator on interest--starter pairs to improve generation quality and coverage;
$\bullet$ \textbf{SFT + DPO} further trains the generator with vanilla direct preference optimization based on user interaction signals.

\begin{table}[t]
\centering
\small 
  \setlength\abovecaptionskip{4pt}
  \setlength\belowcaptionskip{-6pt}
\setlength{\tabcolsep}{1pt} 
\renewcommand{\arraystretch}{1.}
    \begin{tabular}{l cccc}
        \toprule
        \multirow{2}{*}[-0.5ex]{\textbf{Method}} &
        \multicolumn{2}{c}{\textbf{Utility}} &
        \multicolumn{2}{c}{\textbf{Diversity}} \\
        \cmidrule(lr){2-3}\cmidrule(lr){4-5}
        & R-User $\uparrow$
        & R-Score $\uparrow$
        & Lexical $\uparrow$
        & Semantic $\uparrow$ \\
\midrule
    \multicolumn{5}{>{\columncolor{gray!15}}c}{Qwen2.5-7B} \\
    PE   & -- & -- & \textbf{29.03} & \textbf{6.11} \\
    PE + RID    & +0.69 & +0.32 & 24.12 & 4.99 \\
    SFT  & +0.73 & +0.39 & 26.46 & \uline{5.17} \\
    SFT + DPO  & \uline{+0.75} & \uline{+0.42} & 13.75 & 2.64 \\
    \textbf{\modelname}  & \textbf{+0.82} & \textbf{+0.74} & \uline{27.74} & 5.01 \\
\midrule
    \multicolumn{5}{>{\columncolor{gray!15}}c}{Doubao1.5-Lite} \\
    PE   & +0.56 & +0.08 & \textbf{29.45} & \textbf{6.23} \\
    PE + RID   & +0.71 & +0.38 & 25.13 & 4.86 \\
    SFT  & +0.78 & +0.44 & \uline{28.97} & \uline{5.59} \\
    SFT + DPO   & \uline{+0.79} & \uline{+0.52} & 12.94 & 2.37 \\
    \textbf{\modelname}  & \textbf{+0.89} & \textbf{+0.80} & 28.83 & 5.28 \\
\bottomrule
\end{tabular}
\caption{Main offline results across backbones ($\uparrow$ indicates higher is better). Utility reflects relative improvement over PE. The best and second-best results are indicated in \textbf{bold} and \uline{underlined}, respectively. \label{tab:main_offline}}
\end{table}

\subsection{Offline Evaluation and Model Selection} \label{sec:offline_eval}

Offline evaluations compare \modelname against baselines and determine the optimal backbone for deployment. As shown in Table~\ref{tab:main_offline}, we evaluate Qwen2.5-7B and Doubao1.5-Lite across two dimensions: (i) \textbf{Utility} measures alignment with the deployed learning-to-rank model through user-level ranking consistency (fraction of improved users) and score-level lift. 
(ii) \textbf{Diversity} measures diversity using a keyword-based lexical metric and a classifier-based semantic metric. Details are provided in Appendix ~\ref{sec:metric_details}.

Experimental results show that:
(i) \textbf{RID improves personalization.} Distilling resonance-triggering interests provides more user-specific conditioning and improves over direct prompting.
(ii) \textbf{\modelname balances utility and diversity.} Through utility--diversity preference alignment, it achieves the highest ranking consistency without sacrificing variety or collapsing into repetitive outputs.
Doubao-based \modelname is selected for online A/B testing due to its superior performance.

\begin{table}[t]
    \centering
    \renewcommand{\arraystretch}{1.}
    \setlength{\tabcolsep}{4pt}
    \setlength\abovecaptionskip{4pt}
    \setlength\belowcaptionskip{-10pt}
    \small
    \begin{tabular}{l llll}
        \toprule
        \textbf{Method} &
        \textbf{Active} $\uparrow$ &
        \textbf{Avg.S.} $\uparrow$ &
        \textbf{CTR} $\uparrow$ &
        \textbf{CSR} $\uparrow$ \\
        \midrule
        PE & -0.01 & -0.26 & -16.16* & -0.17 \\
        SFT & +0.20 & +0.33 & +6.97* & -0.05 \\
        SFT + DPO & +1.16 & +0.42* & +56.41* & +0.68\\
        \textbf{\modelname} & \textbf{+1.84}* & \textbf{+1.59}* & \textbf{+94.25}* & \textbf{+1.27}* \\
        \bottomrule
    \end{tabular}
    \caption{Main online A/B test results (relative lifts \textperthousand\ over the deployed baseline). We report user active days (Active), average sessions per user (Avg.S.), click-through rate (CTR), and conversation start rate (CSR). The best results are \textbf{bolded} and * indicates $p{<}0.05$. \label{tab:main_online}}
\end{table}

\subsection{Online A/B Tests}

We conduct online A/B tests on one of the world's largest conversational agent products for more than one month. Detailed metric definitions are provided in Appendix~\ref{sec:metric_details}. 
Table~\ref{tab:main_online} shows that:
(i) \textbf{Breaking the first-message barrier.} \modelname achieves the largest CSR lift, indicating that more users cross the first-turn initiation threshold.
(ii) \textbf{Downstream benefits.} Beyond initiation, \modelname consistently improves Active and Avg.S., suggesting sustained engagement after the conversation started.

\subsection{Further Analysis} \label{sec:analysis}

\begin{figure}[t]
  \setlength\abovecaptionskip{4pt}
  \setlength\belowcaptionskip{-9pt}
    \centering
    \includegraphics[width=\linewidth]{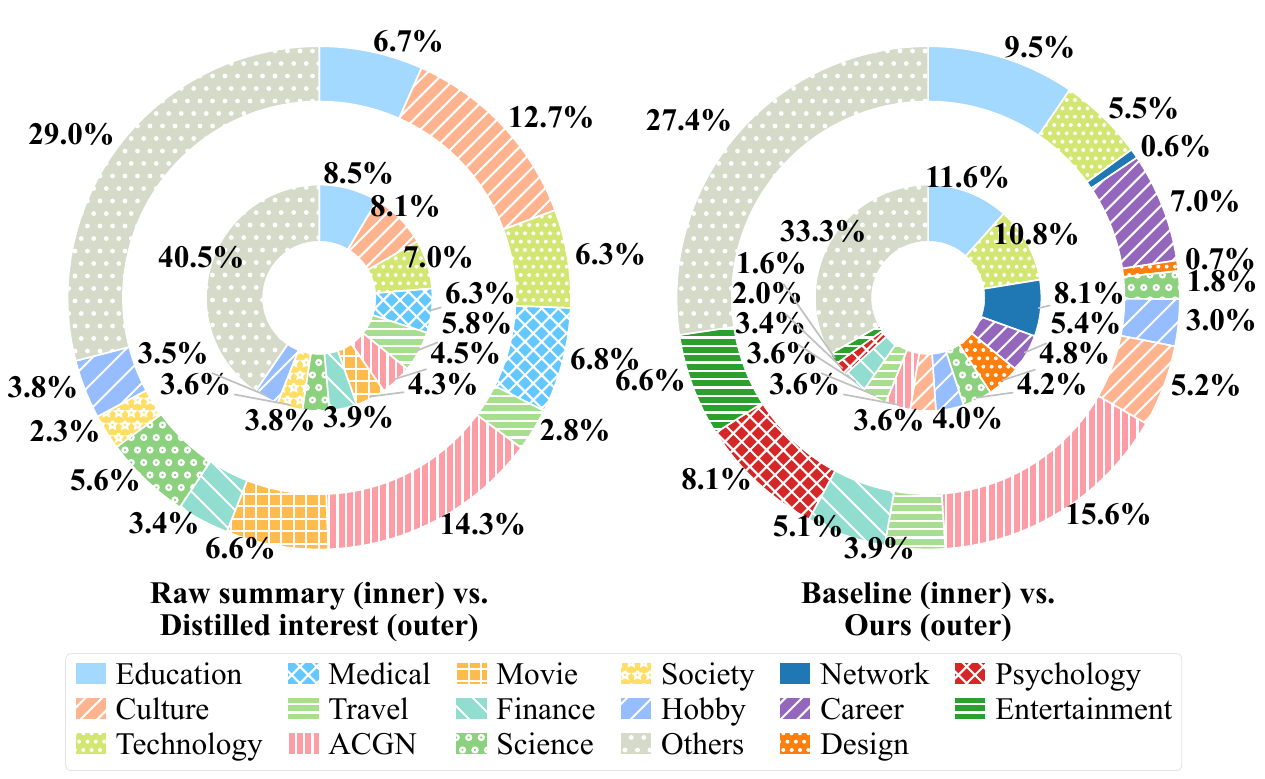}
    \caption{Distribution analysis. \textbf{Left:} RID distills away generic topics toward triggering categories. \textbf{Right:} starters generated by \modelname show broader coverage and higher interaction potential.}
    \label{fig:distill_shift}
\end{figure}

\paragraph{Distribution Analysis}
We map each text to a fine-grained topic category with a pre-trained classifier and visualize topic proportions across pipeline stages.
In \figurename~\ref{fig:distill_shift}, the inner and outer rings show the distributions before and after each step (\textcolor[HTML]{d6dac8}{\rule{1.5ex}{1.5ex}}\,``Others'' aggregates categories below the display threshold):
(i) \textbf{RID denoises and sharpens interests.} Distilled trigger interests downweight functional and generic head topics and shift the distribution toward resonance-prone, long-tailed cues.
(ii) \textbf{ISG steers starters toward ice-breaking topics.} Compared with distilled interests, generated starters tilt further toward easy-to-engage, consumption-oriented categories (e.g., \textcolor[HTML]{dc4848}{\rule{1.5ex}{1.5ex}}\,{``Psychology''}, \textcolor[HTML]{ff9da7}{\rule{1.5ex}{1.5ex}}\,{``ACGN''}, \textcolor[HTML]{56b357}{\rule{1.5ex}{1.5ex}}\,{``Entertainment''}), which are more likely to elicit interaction.
(iii) \textbf{Utility-aligned diversity.} Relative to the deployed baseline, \modelname achieves broader topical coverage with reduced head dominance while remaining aligned with interaction utility, rather than being diverse but irrelevant.

\paragraph{Cases Analysis}
\Tref{tab:case_study} provides an illustrative comparison. Compared with retrieved generic starters, \modelname generates more personalized, interaction-inviting starters. 

\begin{table}[t]
    \centering
    \footnotesize
    \setlength{\tabcolsep}{0pt}
    \renewcommand{\arraystretch}{1.1}
    \setlength\abovecaptionskip{4pt}
    \setlength\belowcaptionskip{-9pt}
    \begin{tabular}{m{7.6cm}}
        \toprule
        \rowcolor{mylightblue}\textbf{Session summaries:} AI trends; Weekend schedule planning; One Piece anime wallpapers; Translating a Latin article on Philosophy; Cosplay beginner tutorial; Top anime opening of 2025; JoJo character analysis; Local LLM deloyment \\
        \midrule
        \textbf{Retrieved conversation starters (Deployed):}\\
        \ding{172} What are the latest LLM papers? \\
        \ding{173} How to spend a fulfilling vacation? \\
        \ding{174} How to better cosplay female anime characters? \\
        \ding{175} Recommend some popular anime openings in 2025. \\
        \midrule
        \rowcolor{mylightblue}\textbf{Distilled trigger interests:} \textit{Anime openings and characters (One Piece, JoJo)}; \textit{cosplay ideas}; \textit{AI trends and applications}\\
        \midrule
        \textbf{Generated conversation starters (\modelname):} \\
        \ding{172} What anime conventions will be held this year? \\
        \ding{173} Which One Piece character is beginner-friendly to cosplay? \\
        \ding{174} Explain techniques for acceleration of LLM deployment.\\
        \ding{175} Three anime movies for a relaxing weekend.\\
        \bottomrule
    \end{tabular}
    \caption{Case study comparing the deployed retrieval-based starters and our \modelname. 
    (1) Denoising: RID filters out functional noise in session summaries (e.g., translation requests) and distills trigger interests. (2) Finer-grained personalization beyond retrieval: compared to generic starters, \modelname generates more personalized starters with finer topical granularity. 
    \label{tab:case_study}}
\end{table}

\section{Related Work} \label{sec:related_work}

Recent advances in LLMs have empowered conversational agents to move beyond passive responding \cite{li-etal-2025-hello, liu2024llm, deng2025proactive}. Prior work on proactivity mainly studies \textit{in-conversation} behaviors, including follow-up question generation \citep{li2025proactive} and topic steering \citep{rebedea2024canttalkaboutthis}, which depend on dialogue context and explicit intents. 
For preference alignment, a growing body of work studies preference alignment for LLMs to better reflect human preferences and improve assistance quality, leveraging feedback signals via supervised fine-tuning and preference optimization \citep{yin2025clickspreferencemultistagealignment, Sun2024RLRF4RecRL, wu-etal-2025-aligning}. However, such general alignment objectives are insufficient for proactive initiation, which requires user-level personalization under sparse feedback. Appendix~\ref{sec:related_work_extended} provides an extended discussion.

\section{Conclusion}
Conversational agents are becoming daily assistants for billions of users.
We highlight proactive initiation as a practical capability for modern LLM chat assistants to overcome the first-message barrier in cold-start moments.
To this end, we study Conversation Starter Generation and propose \textbf{\modelname}, which operationalizes proactive initiation as a two-step handshake: Resonance-Aware Interest Distillation for evoking user resonance and Interaction-Oriented Starter Generation for stimulating interaction.
Extensive offline evaluations and online A/B tests demonstrate that \modelname consistently improves user experiences and conversation quality, providing a scalable solution for industrial-grade proactive conversational agents.

\bibliography{custom}

\appendix

\section{Extended Related Work} \label{sec:related_work_extended}
\subsection{LLM-based Conversational Agents}

Recent advances in LLMs have empowered conversational agents \cite{li-etal-2025-hello, liu2024llm} to move beyond passive responding \citep{deng2025proactive}. Existing works on proactivity primarily focus on \textit{in-conversation} stages, such as generating follow-up questions \citep{li2025proactive, yin2025clickspreferencemultistagealignment}, clarifying ambiguous queries \citep{deng2023prompting}, or steering topics \citep{rebedea2024canttalkaboutthis}. However, these methods rely on ongoing dialogue contexts and explicit user intents, rendering them ineffective for cold-start users who have not yet initiated a conversation. Our work targets the \textit{before-conversation} stage, addressing the critical bottleneck of the first-message barrier.

\subsection{LLM Alignment for User Preference}

Aligning LLMs with user preferences has been extensively studied in recommendation \citep{ZHU2026104434, wang-etal-2025-user, Liao2024RosePOAL}, search \citep{10.1145/3726302.3731955, qin-etal-2025-maps}, and conversational systems \citep{wu-etal-2025-aligning, wang2023aligninglargelanguagemodels}. Recent approaches leverage human feedback, such as clicks or comparative judgments, to align LLMs with user values and interests through Supervised Fine-Tuning (SFT) \citep{10.1145/3726302.3731955}, Reinforcement Learning from Human Feedback  (RLHF) \citep{yin2025clickspreferencemultistagealignment, yang2025rlhffinetuningllmsalignment}, or Direct Preference Optimization (DPO) \citep{Sun2024RLRF4RecRL, Liao2024RosePOAL}. These methods can only align overall platform preferences, not personalized preferences for each user. To address the issue of sparse feedback in personalized preference alignment, we designed a list-wise, multi-objective formulation to bridge the generative decoder with user interests.

\section{More Details of Experiment Setups}

\subsection{Metrics Calculation}
\label{sec:metric_details}
We provide detailed definitions and calculation procedures for the offline and online metrics used throughout our experiments.

\subsubsection{Offline Metrics}

\paragraph{Diversity} measures the coverage of user interests at both lexical and semantic granularities. \textbf{Lexical diversity} employs a TF-IDF-based keyword extraction approach, computing the number of unique high-weight keywords extracted from generated starters. This captures surface-level topical variety through vocabulary distinctiveness.
\textbf{Semantic diversity} employs a classifier-based approach, where we map generated starters to a hierarchical interest taxonomy using a trained topic classifier and compute the number of distinct interest categories activated. This captures deeper semantic variation beyond surface keywords.

\paragraph{Utility}
Utility evaluates the alignment between generated starters and the deployed online learning-to-rank (LTR) model. Let $\mathcal{U}$ denote the set of evaluated users. For each user $u \in \mathcal{U}$, let $\mathcal{G}_u$ and $\mathcal{R}_u$ denote the sets of generative starters and baseline candidates, respectively. $f(\cdot)$ denotes the scoring function of the deployed learning-to-rank model.

\textbf{R-User} measures the fraction of users whose generated starters achieve higher average scores than baseline candidates:

\begin{equation}
\text{R-User} =
\frac{1}{|\mathcal{U}|}
\sum_{u \in \mathcal{U}}
\mathbb{I}
\big(
\bar{f}(\mathcal{G}_u)
>\bar{f}(\mathcal{R}_u)
\big),
\end{equation}

\textbf{R-Score} measures the average LTR score improvement of generative starters over baseline candidates:

\begin{equation}
\text{R-Score} =
\frac{1}{|\mathcal{U}|}
\sum_{u \in \mathcal{U}}
\frac{
\bar{f}(\mathcal{G}_u) - \bar{f}(\mathcal{R}_u)
}{
\bar{f}(\mathcal{R}_u)
},
\end{equation}

\subsubsection{Online Metrics}
For online A/B tests, all metrics are reported as relative lifts over the deployed baseline. Let $\mathcal{U}$ denote the set of users exposed to the system during the evaluation period, and $\mathcal{S}_u$ denote the set of sessions initiated by user $u \in \mathcal{U}$. Let $\mathbb{I}(\cdot)$ be the indicator function.

\textbf{Active} (user active days) measures the number of days a user engages with the system within a 7-day window.

\textbf{Avg.S.} (average sessions per user) measures the average number of conversation sessions per user:
\begin{equation}
\text{Avg.S.} = \frac{1}{|\mathcal{U}|} \sum_{u \in \mathcal{U}} |\mathcal{S}_u|.
\end{equation}

\textbf{CTR} (click-through rate) measures the proportion of recommended starters that receive clicks:
\begin{equation}
\text{CTR} = \frac{N_\text{clicked}}{N_\text{shown}},
\end{equation}

\textbf{CSR} (conversation start rate) measures the proportion of active users who have initiated conversations:
\begin{equation}
\text{CSR} = \frac{|\{u \in \mathcal{U} : \text{start}(u) > 0\}|}{|\mathcal{U}|},
\end{equation}
where $\mathcal{U}$ denotes the set of active users during the evaluation period and $\text{start}(u)$ indicates whether user $u$ starts a conversation, including entering a conversation via clicking a starter.

\subsection{More Implementation Details}
\label{sec:baseline_details}

\subsubsection{Prompts} \label{appendix:prompt}

We summarize the prompts used in our pipeline. Table~\ref{tab:distill_prompt} is used to generate a coverage-oriented interest--starter corpus for Supervised Interest-Expanded Instruction Tuning (SIT), as well as for the prompt-based baseline. Table~\ref{tab:interest_revisit_prompt} is used to detect interest revisits by matching later session summaries to historical ones, providing samples for training the personalized resonance scorer in RID.

\begin{table}[t]
\small
\centering
\begin{tabular}{p{0.45\textwidth}} 
\toprule
\textbf{Task Objective.} \\ 
Based on the information provided, generate several questions and corresponding reasoning processes that \textbf{conform to common sense, share similar interest subjects, are relevant to the input, but do not repeat specific content}. \\
\midrule
\textbf{Key Rules.} 
\begin{enumerate}[leftmargin=*,itemsep=2pt,topsep=0pt,parsep=0pt]
    \item \textbf{Length Constraint}: Each question must be strictly limited to 20 characters;
    \item \textbf{Number of Questions}: Output 5--20 questions;
    \item \textbf{Diversity Requirement}: Cover diverse topic types, no medical topics;
    \item \textbf{Long- and Short-term Interest Balance}: Focus on both core and immediate interests;
    \item \textbf{Generality}: Generalization and association are required;
    \item \textbf{Fine-grained Requirement}: Questions should be specific, not generic;
    \item \textbf{User Expertise Matching}: Adjust professionalism based on user interest;
    \item \textbf{Factuality}: Subjects must objectively exist;
    \item \textbf{Safety}: No sensitive, negative, or invasive topics.
\end{enumerate} \\ 
\midrule
\textbf{Output Format.} \\
\texttt{<think>} Reasoning process \texttt{</think>} \\
\texttt{<sug>} Question 1 \&\& Question 2 \&\& \dots \&\& Question n \texttt{</sug>} \\
\midrule
\textbf{Input.} \\
\{user\_interest\} \\
\bottomrule
\end{tabular}
\caption{Prompt used for distilled data generation. \label{tab:distill_prompt}}
\end{table}

In Table~\ref{tab:distill_prompt}, the input \{user\_interest\} is a set of distilled trigger interests. The model is instructed to output multiple short, diverse, and qualified conversation starters under explicit constraints; we parse the questions from the \texttt{<sug>} field and ignore the intermediate reasoning.

\begin{table}[t]
\small
\centering
\begin{tabular}{p{0.45\textwidth}}
\toprule
\textbf{Task Objective.} \\
Given an earlier session summary and a later session summary from the same user, decide whether the later session revisits the same underlying interest as the earlier one. \\
\midrule
\textbf{Key Rules.}
\begin{enumerate}[leftmargin=*,itemsep=2pt,topsep=0pt,parsep=0pt]
    \item Use only the provided session summaries.
    \item Output \texttt{MATCH} only if both summaries refer to the same specific interest/topic (paraphrases allowed); otherwise output \texttt{NO\_MATCH}.
    \item Shared generic themes (e.g., ``music'', ``travel'') are insufficient for \texttt{MATCH} unless the specific subject aligns.
\end{enumerate} \\
\midrule
\textbf{Output Format.} \\
\texttt{<label>} MATCH or NO\_MATCH \texttt{</label>} \\
\texttt{<rationale>} One-sentence justification \texttt{</rationale>} \\
\midrule
\textbf{Input.} \\
\texttt{[Earlier session summary]}: \{h\_earlier\} \\
\texttt{[Later session summary]}: \{h\_later\} \\
\bottomrule
\end{tabular}
\caption{Prompt used for interest-revisit detection via session-summary matching. \label{tab:interest_revisit_prompt}}
\end{table}

In Table~\ref{tab:interest_revisit_prompt}, we apply the prompt to pairs of historical and later session summaries from the same user. We treat \texttt{MATCH} pairs as interest-revisit positives for RID training and discard the optional \texttt{<rationale>} in downstream processing.

\subsubsection{Implementation Details} \label{appendix:implementation}

\paragraph{Model Architecture}
We experiment with two backbone models as the base generator $\pi_{\theta}$:
\begin{itemize}[leftmargin=*,itemsep=1pt,topsep=2pt,parsep=0pt]
    \item \textbf{Qwen2.5-7B}: A 7-billion-parameter open-source model with 32 transformer layers, 4096 hidden dimensions, and 32 attention heads. The model supports a context window of 32K tokens.
    \item \textbf{Doubao1.5-Lite}: A proprietary lightweight model optimized for low-latency inference with comparable capacity. The model employs efficient attention mechanisms and is trained with multi-task instruction tuning. Model details are available at \url{https://console.volcengine.com/ark}.
\end{itemize}

\paragraph{Deployment Latency}
In production deployment, \modelname meets real-world latency requirements: the median end-to-end latency for generating a full starter list is 2.9 seconds, and the first-token latency is 0.28 seconds. \modelname has been deployed at scale in the product.

\section{Additional Experimental Results} \label{sec:more_rets}

We conduct extensive studies to quantify the contribution of key design choices in our system.
All experiments are evaluated using the same offline metrics as in Section \ref{sec:offline_eval}. The results validate that our method designs are complementary and jointly necessary to improve both utility and diversity.

\subsection{Analysis on Resonance-Aware Interest Distillation}

The RID component is designed to distill high-resonance trigger interests. We further investigate its effectiveness through a two-step evaluation: (1) a direct quality assessment of the distilled interests, and (2) their downstream impact on starter generation. We compare RID against two baselines: (i) \textbf{Non-Distillation}, which feeds all historical interests into the generator without filtering; and (ii) \textbf{Rule-based}, which selects interests based on static heuristics such as frequency of occurrence and recency within the user's history summary.

\begin{table*}[t]
    \centering
    \small
    \renewcommand{\arraystretch}{1.15}
    \setlength{\tabcolsep}{2.5pt}
    \begin{tabular}{lccccccc}
        \toprule
        \multirow{2}{*}[-0.5ex]{
            \begin{tabular}{c}
            \textbf{Distillation} \\
            \textbf{Strategy}
            \end{tabular}
            } &
        \multicolumn{3}{c}{\textbf{Interest Quality}} &
        \multicolumn{2}{c}{\textbf{Utility}} &
        \multicolumn{2}{c}{\textbf{Diversity}} \\
        \cmidrule(lr){2-4}\cmidrule(lr){5-6}\cmidrule(lr){7-8}
        & DeepSeek-R1 & Doubao-1.8 & Human
        & R-User $\uparrow$ & R-Score $\uparrow$ 
        & Lexical $\uparrow$ & Semantic $\uparrow$ \\
        \midrule
        Non-Distillation & -- & -- & -- & -- & -- & \textbf{29.45} & \textbf{6.23}  \\
        Rule-based & 0.07 & 0.08 & 0.14 & +0.70 & +0.35 & 25.98 & 4.87  \\
        Reward-based (RID) & \textbf{0.93} & \textbf{0.92} & \textbf{0.86} & \textbf{+0.78} & \textbf{+0.58} & 28.83 & 5.28  \\
        \bottomrule
    \end{tabular}
    \caption{Evaluation of interest distillation strategies. \textbf{Interest Quality} reports the pairwise win rate from LLM-as-judge (DeepSeek-R1-671B and Doubao-Seed-1.8-High) and human annotators. \textbf{Utility} and \textbf{Diversity} evaluate the downstream generation quality. \label{tab:rid_evaluation}}    
\end{table*}

\paragraph{Interest Quality Evaluation}
To quantify the quality of distilled interests directly, we employ both LLM-as-judge and human evaluations in a pairwise comparison setting. 
Given a user's raw interests (Non-Distillation), both LLMs (DeepSeek-R1-671B and Doubao-Seed-1.8-High) and human annotators are asked to judge which distillation strategy produces interests that are more likely to serve as high-resonance triggers for a future conversation.
As shown in Table~\ref{tab:rid_evaluation}, \textbf{RID achieves a dominant win rate over rule-based distillation across all evaluators}. This confirms that static heuristics often surface generic or repetitive interests, while RID accurately identifies specific, high-reward points aligned with user resonance.

\paragraph{Effect on Starters Generation}
We further investigate how the quality of distilled interests translates into final generation performance using the offline evaluation metrics described in \S\ref{sec:offline_eval}. The non-distillation setting, while having the highest diversity due to its broad input, underperforms at utility, indicating that a lack of distillation introduces significant noise. Rule-based retrieval improves utility but is still outperformed by RID. Our \textbf{reward-based strategy (RID)  achieves the best utility while maintaining good diversity}, demonstrating that by accurately distilling high-resonance seeds, RID effectively guides the generator to produce conversation starters that are both personalized and engaging.

\subsection{Analysis on Self-Reinforced Iterative Optimization} \label{appendix:temporal}

During the self-reinforced iterative optimization process, the model is continually updated using time-sliced data to harvest the latest supervision signals. Each slice corresponds to user interests collected within a fixed window (e.g., one week) before a specific date, paired with candidates scored by the ranking model deployed at that time. During evaluation, all models are assessed using the latest ranking model and the same user inputs, ensuring that performance reflects adaptability to the current online environment.

\begin{table}[t]
    \centering
    \small
    \renewcommand{\arraystretch}{1.1}
    \setlength{\tabcolsep}{1pt}
        \begin{tabular}{lcccc}
        \toprule
        \multirow{2}{*}[-0.5ex]{
            \textbf{Data Slice}
            } &
        \multicolumn{2}{c}{\textbf{Utility}} &
        \multicolumn{2}{c}{\textbf{Diversity}} \\
        \cmidrule(lr){2-3}\cmidrule(lr){4-5}
        & R-User $\uparrow$
        & R-Score $\uparrow$ 
        & Lexical $\uparrow$
        & Semantic $\uparrow$ \\
        \midrule
        Early  & +0.79 & +0.40 & 30.76 & \textbf{6.25} \\
        Middle & +0.85 & +0.70 & \textbf{30.82} & 6.14  \\
        Latest & \textbf{+0.89} & \textbf{+0.80} & 28.83 & 5.28  \\
        \bottomrule
    \end{tabular}
    \caption{Effect of Self-Reinforced Iterative Optimization using different training periods. All models are evaluated with the latest ranking model and compared against the prompt engineering model as the baseline.}
    \label{tab:period_alignment}
\end{table}

As shown in Table \ref{tab:period_alignment}, \textbf{generators trained on more recent data slices consistently achieve higher utility}, confirming that iterative optimization enables the generator to better adapt to evolving ranking preferences and user behavior. However, we also observe a gradual decrease in diversity as the training slice becomes more recent. We attribute this phenomenon to the increased concentration of supervision signals from the current ranking model, which tends to emphasize a narrower set of high-reward interests.

\subsection{Analysis on Training Data Usage}

To understand how training data usage affects generation quality, we ablate the data composition in both the SIT and PMA stages while keeping the backbone and evaluation protocol unchanged.
(i) For SIT, we compare our distilled-only setting that instruction-tunes the generator on teacher-distilled interest--starter pairs (\modelname) against a variant that additionally mixes in self-reinforced data (\textbf{w/ Self-Reinforced}). (ii) For PMA, we compare our default preference-pair construction that mixes distilled starters with self-reinforced candidates (\modelname) against removing the distilled starters (\textbf{w/o Distilled}). 

Results shown in Table~\ref{tab:ablation_data_composition} validate our design: SIT should prioritize coverage expansion through teacher distillation to establish a broad sample space and raise the model's capability ceiling, while PMA refines utility-diversity trade-offs through personalized preference learning.

\begin{table}[t]
    \centering
    \small
    \setlength{\tabcolsep}{0pt} 
    \renewcommand{\arraystretch}{1.1}
    \begin{tabular}{lcccc}
        \toprule
        \multirow{2}{*}[-0.5ex]{
            \begin{tabular}{c}
            \textbf{Data} \\
            \textbf{Source}
            \end{tabular}
            } &
        \multicolumn{2}{c}{\textbf{Utility}} &
        \multicolumn{2}{c}{\textbf{Diversity}} \\
        \cmidrule(lr){2-3}\cmidrule(lr){4-5}
        & R-User $\uparrow$
        & R-Score $\uparrow$
        & Lexical $\uparrow$
        & Semantic $\uparrow$ \\
        \midrule
        \multicolumn{5}{>{\columncolor{gray!10}}c}{SIT Data Composition}\\
        \textbf{\modelname} & \textbf{+0.89} & \textbf{+0.80} & \textbf{28.83} & \textbf{5.28} \\
        \; w/ Self-Reinforced & +0.83 & +0.32 & 19.6 & 4.79  \\
        \midrule
        \multicolumn{5}{>{\columncolor{gray!10}}c}{PMA Data Composition} \\
        \textbf{\modelname} & \textbf{+0.89} & \textbf{+0.80} & \textbf{28.83} & \textbf{5.28} \\
        \; w/o Distilled & +0.85 & +0.34 & 16.12 & 3.20  \\
        \bottomrule
    \end{tabular}
    \caption{Analysis of training data composition for SIT and PMA stages. The best results of each stage are indicated in \textbf{bold}.}
    \label{tab:ablation_data_composition}
\end{table}

\paragraph{SIT Data Composition} 
Using distilled supervision for SIT yields the best performance. In contrast, adding self-reinforced data at the SIT stage (w/ Self-Reinforced) degrades both utility and diversity, suggesting that early instruction tuning benefits more from clean, high-coverage distilled signals than from on-policy self-generated data.

\paragraph{PMA Data Composition}
For PMA, mixing distilled starters with self-reinforced candidates achieves the best results. Removing distilled starters (w/o Distilled) leads to clear drops in both utility and diversity, indicating that distilled data provides complementary coverage that improves preference learning and avoids collapsing to a narrow set of repetitive patterns.

\subsection{Analysis on Hierarchical Negative Sampling}

\begin{table}[t]
    \centering
    \small
    \setlength{\tabcolsep}{1pt}
    \renewcommand{\arraystretch}{1.1}
        \begin{tabular}{lcccc}
        \toprule
        \multirow{2}{*}[-0.5ex]{
            \begin{tabular}{c}
            \textbf{Negative} \\
            \textbf{Sampling}
            \end{tabular}
            } &
        \multicolumn{2}{c}{\textbf{Utility}} &
        \multicolumn{2}{c}{\textbf{Diversity}} \\
        \cmidrule(lr){2-3}\cmidrule(lr){4-5}
        & R-User $\uparrow$
        & R-Score $\uparrow$ 
        & Lexical $\uparrow$
        & Semantic $\uparrow$ \\
        \midrule
        \textbf{Utility-Only} 
        & \textbf{+0.85} 
        & \textbf{+0.36}
        & 15.18
        & 2.94 \\
        \textbf{Diversity-Only} 
        & +0.58 
        & +0.08 
        & \textbf{33.34}
        & \textbf{6.50} \\
        \textbf{Joint-failure} 
         & +0.72 
        & +0.20
        & 21.3
        & 4.59 \\ 
        \textbf{Hierarchical}
        & \underline{+0.84} 
        & \underline{+0.35}
        & \underline{28.83}
        & \underline{5.28} \\
        \bottomrule 
    \end{tabular}
    \caption{Ablation study of hierarchical negative sampling strategies in the PMA stage. All results are evaluated using the same ranking model and compared against the PE baseline. The best result is shown in \textbf{bold} and the second best is \uline{underlined}. \label{tab:hierarchical-negatives}}
\end{table}

We ablate the hierarchical negative sampling design in the DPO stage by selectively using different types of negative preference pairs. As shown in Table~\ref{tab:hierarchical-negatives}, only using utility negatives improves utility but significantly harms diversity, while diversity negatives enhance diversity with limited gains in ranking-related metrics. Simpler negatives like joint-failure negatives provide weak but stable supervision across all dimensions. In contrast, the hierarchical combination of all negative types achieves the most balanced performance, indicating that structured negative sampling is crucial for jointly optimizing engagement and diversity.

\section{Limitations}
\label{sec:limitations}

While \modelname demonstrates strong performance in generating personalized conversation starters, a few limitations remain. (1) \textbf{Hallucination}: Like other generative models, \modelname may occasionally produce starters that contain factually incorrect information or hallucinated content. In deployment, we mitigate this risk through safety classifiers that detect potentially problematic content. However, fully eliminating hallucination remains an open challenge, particularly for long-tail interests where training supervision is sparse. (2) \textbf{Cold-Start for New Users}:
The Resonance-Aware Interest Distillation (RID) component requires historical session summaries to identify trigger interests effectively. Future work could explore incorporating zero-shot or few-shot learning techniques to better handle cold-start scenarios.

\end{document}